\documentclass{article}

\usepackage{arxiv}

\usepackage[utf8]{inputenc} 
\usepackage[T1]{fontenc}    
\usepackage{hyperref}       
\usepackage{url}            
\usepackage{booktabs}       
\usepackage{amsfonts}       
\usepackage{nicefrac}       
\usepackage{microtype}      
\usepackage{lipsum}		

\usepackage{tikz}
\usepackage{tikzscale}

\usepackage{graphicx}
\usepackage{subfigure}
\usepackage{multirow}
\date{} 
\title{RepGN:Object Detection with Relational Proposal Graph Network}

\author{
  Xingjian Du \\
   \And
  Xuan Shi\\
  \And 
  Risheng Huang
}

\begin{document}
\maketitle
\begin{abstract}
\label{sec:abs}
Region based object detectors achieve the state-of-the-art performance, but few consider to model the relation of proposals.
Most object detection frameworks rely on recognizing and localizing object instances individually. 
In this paper, we explore the idea of modeling the relationships among the proposals for object detection from the graph learning perspective. \\
Specifically, we present relational proposal graph network (RepGN) which is defined on object proposals and the semantic and spatial relation modeled as the edge.
By integrating our RepGN module into object detectors, the relation and context constraints will be introduced to the feature extraction of regions and bounding boxes regression and classification.
Besides, we propose a novel graph-cut based pooling layer for hierarchical coarsening of the graph, which empowers the RepGN module to exploit the inter-regional correlation and scene description in a hierarchical manner.
We perform extensive experiments on COCO object detection dataset and show promising results.
\end{abstract}
\keywords{object detection, proposal, relation, graph convolution network}

\section{Introduction}
\label{sec:int}
 Object detection is one of the most highlighted problems in computer vision. 
 Many real-world applications, including image retrieval, advanced driver assistance system and video surveillance, are based on high-performance object detection.
 In the past decades, impressive achievements have been made to improve the performance of object detection in both accuracy and speed. 
 In the most of prevalent object detection baselines \cite{faster-rcnn,cascade-rcnn} exploiting the feature of proposals individually still the primary choice to improve performance instead of taking correlation information among proposals into account, although it has been a universal concept that context information helps image understanding \cite{chen2017spatial} and relationship among proposals is widely applied in many other computer vision tasks \cite{jiayajia}.
 
 Analysing the several frameworks leveraging the correlation among proposals into object detection \cite{han2018relation}, we notice several reasons why these methods are not popular:
\begin{enumerate}
    \item 
    extra models are  needed for the context information, resulting to the decrease of time and memory efficiency
    \item 
    it is difficult to model the complex relationship among proposals properly, including modelling the spatial and semantic, local and global information simultaneously.
\end{enumerate}

To circumvents the restraints mentioned above, we investigate to depict the relationship among proposals by leveraging the graph convolutional neural network(GCN)\cite{bruna2014spectral}, which seems a more reasonable model than others\cite{jiayajia} \cite{hu2018relation}.
 Proposals in one image are supposed as the vertexes and the Intersection over Union(IoU) is weight of edges between proposal vertexes.
Thanks to the low computation complexity of GCN, our model makes an efficient and effective improvement in AP performance without significant drop in time consumption.\par

Another contribution of this work is we propose graph cut layer to introduce global feature to each proposal.Inspired by the successful application of non-local information in object detection\cite{luo2017non-local}, we attempt to extract global feature by the graph.Specifically, we coarsen the graph hierarchically to get a sparse and global description of all proposals by a pooling methods based on graph cut algorithm.\par

We evaluate our model on COCO dataset and achieve the stable improvement compared with baseline. 
Our method and experiment result are demonstrated in the following part of this paper.
\section{Related Works}
\label{sec:related}

\textbf{Object Detetcion}

Before CNN is prevalent in Object Detection, DPM\cite{DBLP:journals/pami/FelzenszwalbGMR10} is a widely studied model detecting objects by sliding windows in image pyramids.
R-CNN\cite{DBLP:conf/cvpr/GirshickDDM14} makes great success by applying CNNs to process object proposals.
Moreover, calculating feature maps globally instead of locally and sharing computation, Fast R-CNN\cite{fast-rcnn} and SPPNet\cite{DBLP:conf/eccv/HeZR014} enhance the performance of the model.
Faster R-CNN\cite{faster-rcnn} further develops a network that can generate object proposal of high quality, namely region proposal network, which increases the calculating speed to a greater degree.
There are also other methods proposed recently\cite{DBLP:conf/nips/DaiLHS16, DBLP:conf/iccv/DaiQXLZHW17, DBLP:conf/cvpr/LiuQQSJ18} to improve object detection performance by altering network structures.

\noindent{\textbf{Relation and Contextual Information}}

Many object detection methods basing deep neural networks only use internal features to classify the proposals. 
However, the relation with surroundings and contextual information are also important for object detection. 
Bell et al. \cite{bell2016inside} adopted skip pooling and spatial recurrent neural networks to construct a Inside-Outside Net, which integrated both the inside and outside information of the regions of interest.  
Li et al. \cite{li2017attentive} firstly constructed an attention-based global contextualized sub-network that adopted multi-layer long short-term memory networks to generate an attention map representing the global contextual information, and a multi-scale local contextualized sub-network to capture surrounding local context, then imported the global and local contextual information into the region-based object detectors. Ma et al. \cite{ma2017yesnet} developed feature extractor by packing the recurrent neural networks into the convolutional neural networks, with which it can combine global and local information in object detection. 
All these efforts tried to extend the features with local or global contextual information, which demonstrate the importance of relation and contextual information in object detection.

\noindent{\textbf{Graph Neural Network}}

Graph neural networks, or GNNs, are deep learning based methods that operate on graph domain\cite{Jiezhou2018review}. 
Early studies on graph neural network done by Gori et al. and Scarselli et al.\cite{gori2005new,scarselli2009graph} adpoted recurrent network architectures which can transfer information between neighbor nodes to learn a new representation of the target node.

\begin{figure}[h]
	\centering
	\includegraphics[width=0.5\linewidth]{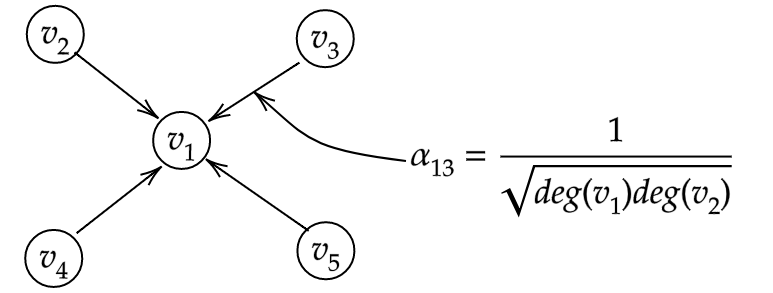}
	\caption{A non-parametric weight $\alpha _{13}$ is assigned  to $v_i$'s neighbor $v_j$ in GCNs.}
	\label{1}
\end{figure}

Inspired by the convolutional neural networks, Bruna et al.\cite{bruna2014spectral} defined graph convolution based on spectral graph theory and applied it in node classification. Further studies such as Edwards et al. \cite{Edwards:2016vy} and Defferrard et al.\cite{defferrard2016convolutional} also followed to make improvements and extensions on spectral based graph convolutional networks. 
Duvenaud et al.\cite{duvenaud2015convolution}, Atwood et al.\cite{atwood2016diffusion} and Hamilton et al.\cite{hamilton2017inductive} define convolutions directly on graph, namely non-spectral based graph convolutional networks, which operate on neighbors spatially. One kind of GCNs is illustrated in Fig.\ref{1}.

A Graph is $G=( V,\ E,\ A)$ where $V$ is the set of nodes, $E$ is the set of edges, and $A$ is the adjacency matrix. In a graph, let $\displaystyle v_{i} \in V$ to denote a node and $\displaystyle e_{ij} =( v_{i} ,\ v_{j}) \in E$ to denote an edge. The adjacency matrix $\displaystyle A$ is a $\displaystyle N\times N$ matrix with $\displaystyle A_{ij} =w_{ij}  >0$ if $\displaystyle e_{ij} \in E$ and $\displaystyle A_{ij} =0$ if $\displaystyle e_{ij} \notin E$. The degree of a node is the number of edges connected to it. For each node $\displaystyle v_{i}$ in the graph $\displaystyle G$, the graph convolution can be formulated as learning a function $\displaystyle f$ that takes node $\displaystyle v_{i}$'s feature $\displaystyle X_{i}$ and $\displaystyle v_{i}$'s neighbors' features $\displaystyle X_{j}$ as input, outputs $\displaystyle v_{i}$'s new representation, where $\displaystyle j\in Neighbor( v_{i})$. Many other graph neural networks takes graph convolution as its core part, such as auto-encoder based models\cite{kipf2016autoencoder} and spatial-temporal networks\cite{bingyu2018spatiotemporal}, etc. 

Based on attention mechanism, Petar et al. \cite{petar2018gat} developed graph attention networks, or GATs, which are actually a spatial-based graph convolution network. The key difference in GATs is that GATs assign more weight to more important nodes by involving attention mechanism which is learned together in model, as illustrated in figure 2. 

\begin{figure}[h]
	\centering
	\includegraphics[width=0.5\linewidth]{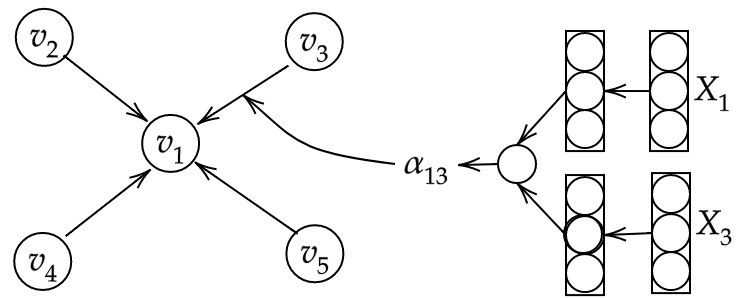}
	\caption{GATs trains a neural network to learn the weight $\alpha_{ij}$, so that larger weight is assigned to more important node.}
	\label{fig:mcmthesis-logo}
\end{figure}

There are also other beautiful efforts on GNNs, for example, Henaff et al.\cite{henaff2015deep} extended the graph convolutional networks to large scale datasets like ImageNet Object Recognition, text categorization, and bioinformatics. Meanwhile, Niepert et al.\cite{niepert2016learning} proposed an approach of PATCHY-SAN, which defined operations of node sequence selection, neighborhood assembly, and graph normalization. As we will show later, these models successfully made CNN work under the graph settings, but they still lack of careful considerations for the specialties of the graph structures in the network design.

\section{Our Approach}
In the object detection task, the correlation between proposals always exist, and they deserve to be exploited better to enrich the representation of proposals. However, there are two types of difficulties in modelling the correlation of proposals. First, the structure of proposals and relations among them are not well defined. Besides, the relational and contextual information extraction module should be efficient.\par

\begin{figure*}[h]
    \centering
    \includegraphics[width=0.95\textwidth]{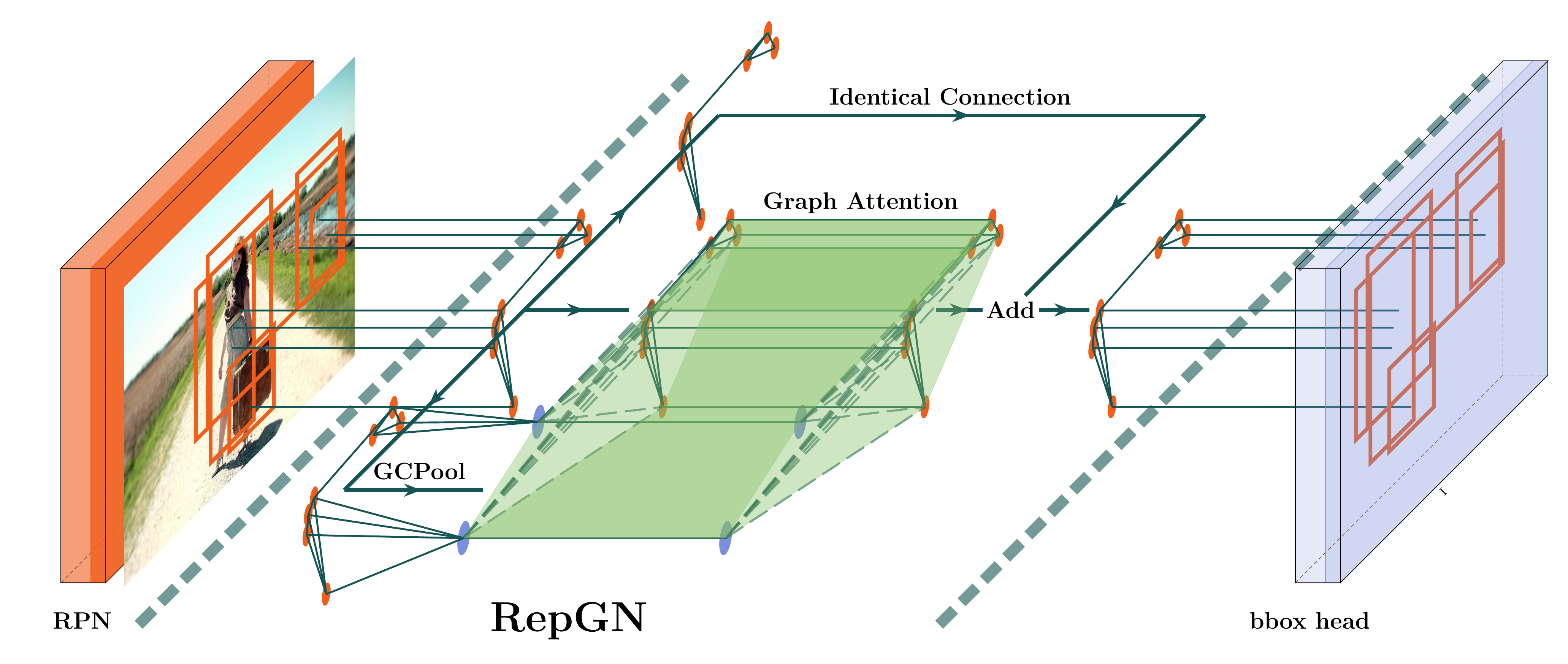}

    \caption{Depiction of our proposed RepGN for object detection. The RepGN consists of the graph attention module, graph cut pooling(GCPool) and the identical connection. The input of the RepGN are the sets of $M$ region proposals from region proposal network(RPN) in the typical object detection pipeline. Then we build the graph of proposals based on the spatial location and intersection over proposals. Besides, we use GCPool to generate to coarsen proposal graph. The coarsen version of proposals graph contains the global contextual information of the input image. To exploit this global information, the sparsity version of proposal graph is combined into the origin proposal. 
    After the construction of the graph, each proposal and its semantic representation is treated as a node in the graph, and a graph attention module exploits the correlation of the vertex. The output of the graph attention module is $M$ refined, local and global contextual aware representation of every original proposal.}
    \label{arch}
\end{figure*}



In this section, we propose the relational proposal graph network (RepGN) to leverage the coherent relationship between the objects in terms of both spatial and co-occurrence probability occurring naturally in the real world.\cite{zhu2017deep}. 
The utilization of spatial and semantic relation helps the estimation of the class and location of an object. 
We build a graph convolution based module to refine the estimation of location and the class probability of proposals from object detector by performing the spatial and semantic reasoning. 
Specifically, this module takes all object proposals generated by regional proposal network as nodes in an undirected simple graph and create edges between every two proposals which have overlap. 
The following parts illustrates the pipeline of object detection we proposed in detail.\par


\subsection{RepGN}
\begin{figure*}[h]
    \centering
    \includegraphics[width=0.95\textwidth]{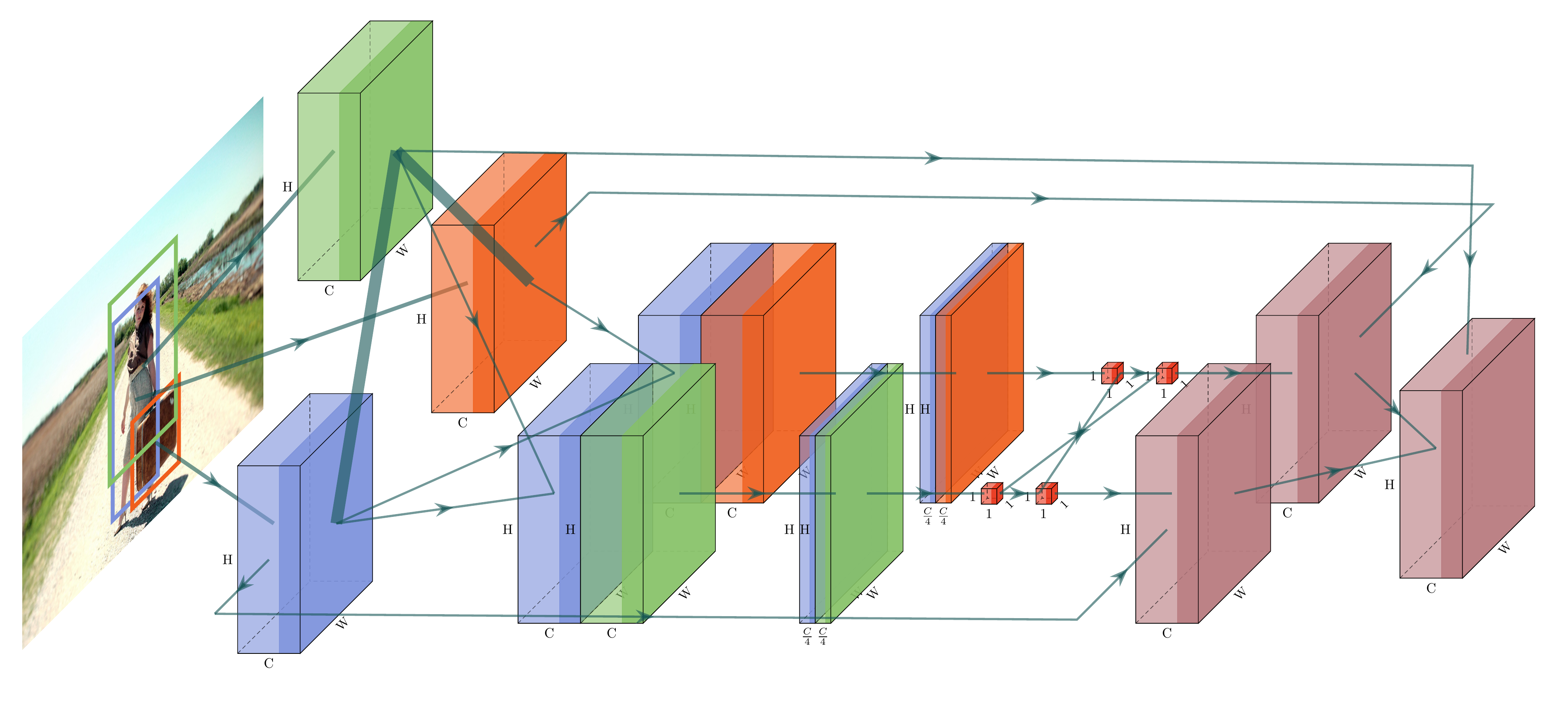}
    \caption{The graph attention module is applied to the proposal graph. For any node, we concatenate its feature matrix to its neighbourhoods on the graph. Then we apply a convolution and average pool operation to this set of concatenated feature matrix to produce semantic similarity between the node and every neighbourhood. Finally, we weight and sum the representation of all neighborhouds depends on their semantic similarity score.}
    \label{gat}
\end{figure*}
In the typical pipeline of object detection, detectors or region proposal networks is firstly leveraged to produce a set of detected objects or regions of interest.
As is shown in Figure \ref{arch} , features of the RoI generated with RPN are feed into the RepGN.
Due to ROI Pooling\cite{faster-rcnn} or ROI Align\cite{faster-rcnn}, shape of RoI's feature map can be processed to uniform and we then treat these features as nodes.
In formalization, we define the input of the RepGN, \emph{i.e.} the output of RPN as features $\mathbf{V}_{i} \in \mathbb{R}^{H\times W\times C}$ and positions $\mathbf{P}_{i} \in \mathbb{R}^{4}$, here $i$ ranges from 1 to $M$, represents the index of ROI. $H$, $W$ and $C$ are the height, width and the number of channels of feature $\mathbf{V}_{i}$ respectively. 
Furthermore, by treating each image region $v_i$ as one vertex, we can construct proposal graph $\mathcal{G}=(\mathcal{V},\mathcal{A})$, where $\mathcal{A}$ denotes the set of the spatial relation between region vertices. In this work, we treat the Intersection Over Union(IoU) as the weight of edges between proposals vertices.\par
After the construction of proposals graph on the spatial domain, each vertex ${v}_{i} \in \mathcal{V}$ is represented by a multi-dimensional vector with its dimension of $1\times(H\times W\times C)$. 
As is shown in Fig. \ref{gat}, for each node, we concatenate its feature matrix to its neighbourhoods on the graph. 
Then we apply a convolution and average pool operation to this set of concatenated feature matrix to produce semantic similarity between the node and every neighbourhood.  
The larger the value is, the higher similarity and the stronger correlation will be. 
The convolution operation gives us an $M$ by $M$ affinity matrix $\mathcal{A}$.\par
In this matrix $\mathcal{A}$, each row $\mathcal{A}_{i}$ represents the correlation between the corresponding node and all the nodes, including itself. 
$\mathcal{A}_{i}$ is then given to softmax to normalize. 
We then compute the multiplication of the normalized attention and the original $M$ nodes, that is, weight and add all nodes according to the correlation weight, and finally get the refined vertex $\mathbf{v}^{\prime}$ which integrates all nodes information.\par
\begin{equation}
\mathbf{v'}_{i}=Softmax(\mathbf{A}_{i})\cdot \mathbf{v},\qquad \mathbf{A}_{i}\in \mathbb{R}^{1\times M}
\end{equation}

Thus far, we retrieved the relation-aware  representation for every proposal by a graph attention module. But for the computation efficiency of our method, we can not afford to stack multiple graph attention layers to expand the receptive field of every node. In this limitation, we can not get non-local and global contextual information through solely one graph attention layer. This problem provides a motivation for us to propose a new graph pooling method for extract the global information in a hierarchy way. This graph croasen method is based on normalized cut of graph which will be introduced on next part.  
\subsection{Graph Cut Pool}
In this section, we propose a type of new pooling method aiming to get a general feature for the ground truth and help to diminish the local semantic feature difference for proposals. 

Nevertheless, some factors hinder this assumption, $1)$ unbalanced proposals, $2)$ densely distributed ground truth. 
Experiment in \cite{jiayajia} have demonstrated that the majority of negative proposals have little effect on improving performance, so it is necessary to filter easy isolated negative proposal candidates at the first time.
Also, proposals for gathered ground truths always distribute densely and are difficult to separate.

\subsubsection{Normalized Cut}
Normalized Cut(NCut)\cite{norm_cut} is a classical graph cut algorithm and designed to circumvent cutting small sets of isolated nodes in the graph by computing fraction of the total edge connections to all the nodes in the graph.

Given a graph $\mathcal{G}=(\mathcal{V}, \mathcal{A})$, NCut aims to divide it into $k$ disjoint sets to maximize weights of each sub-graph $A_{j}$ and minimize the weights of \textit{cut}. 

\begin{equation}
     N_{cut} = \sum_{j=0}^{k} \frac{cut(A_{j}, V/A_{j})}{assoc(A_{j}, V)}
\end{equation}

where the 
$cut(X, V) = \sum_{u \in X, v \in V} {w(u,v)}$ 
and 
$assoc(X, V) = \sum_{u \in X, t \in V} {w(u,t)}$, and $assoc(X,V)$ means the total connection from graph set $X$ to $V$. 

By minimizing the cost of $N_{cut}$, $\mathcal{G}$ gets balanced partitioned and little bias. 
More details can be referred from \cite{norm_cut}.

\subsubsection{Graph Cut Pool}
To eliminate the restrictions mentioned at the beginning of section, we adopt a two-step graph cut pooling approach.
Considering the proposal candidates in one image compose a graph, proposals are vertexes whereas adjoint matrix is $IoU > N_{IoU\_thr}$. 
On the first stage, we remove the connected components whose number of vertices is less than $N_{num\_thr}$ to refine the proposal candidates. 
Fig.\ref{graph_cut} (b) shows the connected components filtered. 

On the second stage, we resort to classical normalized cut algorithm \cite{norm_cut} on each connected components hoping separate proposals in one connected component into partitions. 
Fig.\ref{graph_cut} (d) shows the separated partitions by normalized cut and isolated node sets are filtered again according to step one.

After conducting graph cut hierarchically and iteratively, proposals are divided into several part and assigned pseudo labels.
Proposals with same pseudo labels comprise the $G_{j}(V_{j}, A_{j})$.
On the last but most primary stage, we conduct average graph pooling on each graph. 

\begin{equation}
    v'_{j} = \frac{1}{n}\sum_{v_{jk} \in V{j}} v_{jk} 
\label{xiaji}
\end{equation}

In Eq. \ref{xiaji}, $n$ is the number of vertex in $G_{j}$ and $v'_{j}$ is the output of the Graph Pool Layer.

\def\subw{0.24}
\begin{figure*}
    \centering
    \subfigure[]{
    \includegraphics[width=\subw\textwidth]{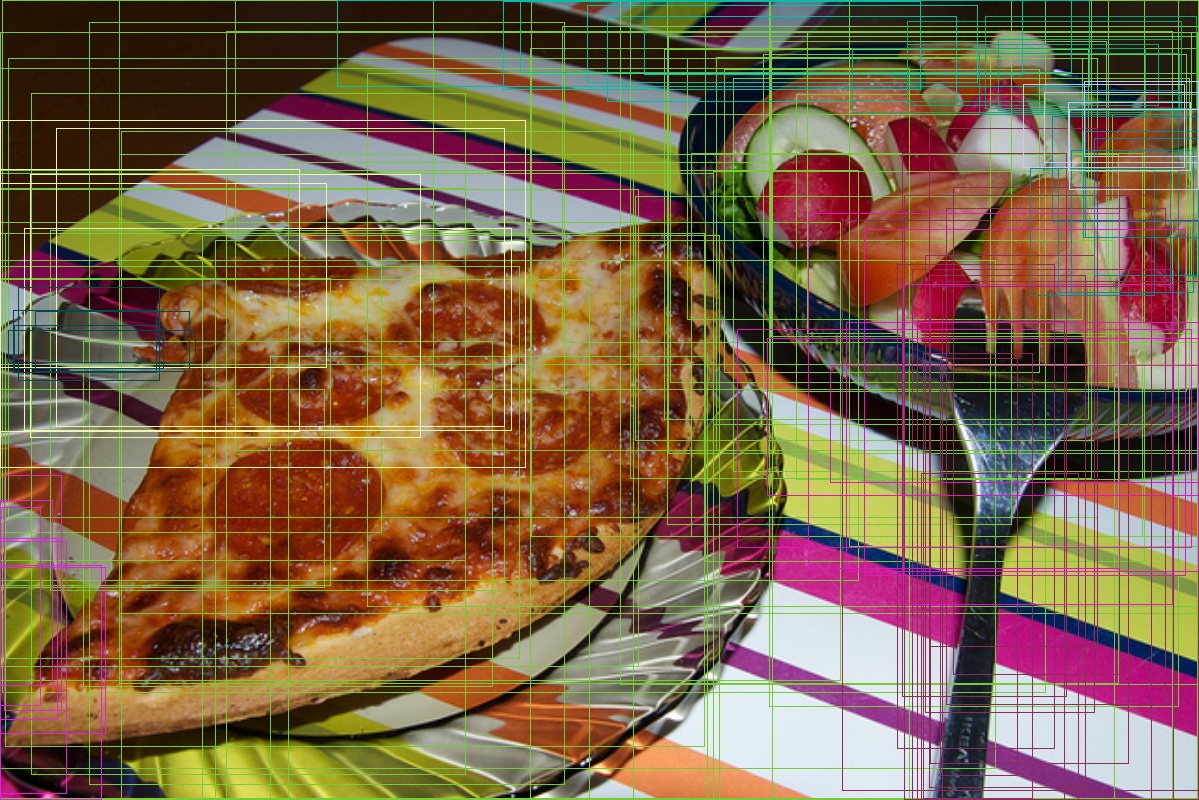}
    }
    \quad
    \subfigure[]{
    \includegraphics[width=\subw\textwidth]{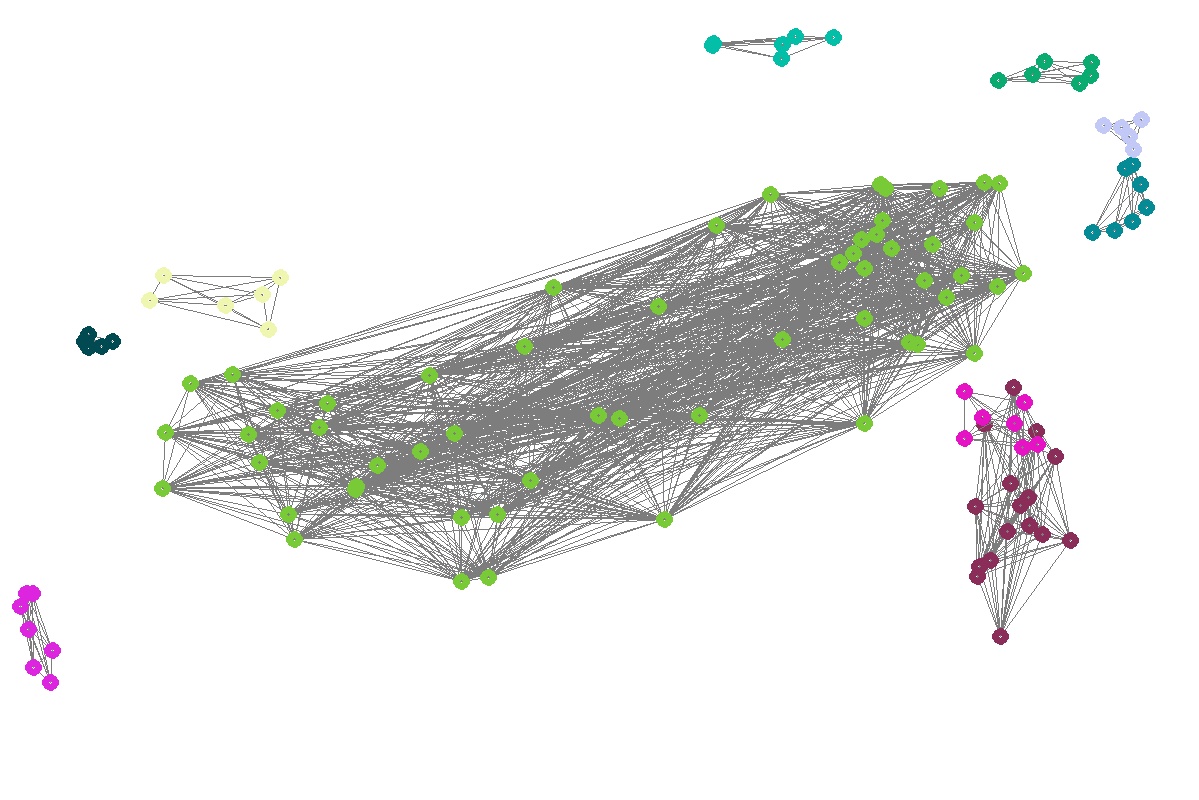}
    }
    \quad
    \subfigure[]{
    \includegraphics[width=\subw\textwidth]{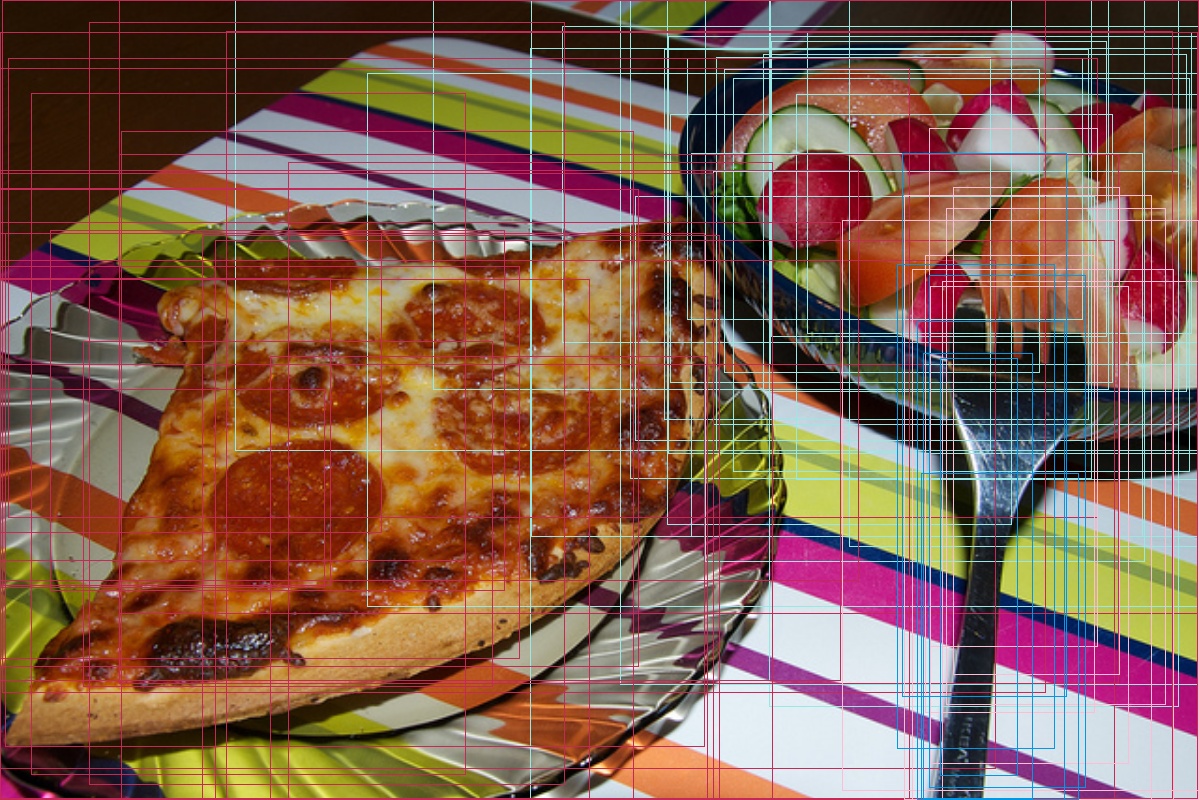}
    }
    \quad
    \subfigure[]{
    \includegraphics[width=\subw\textwidth]{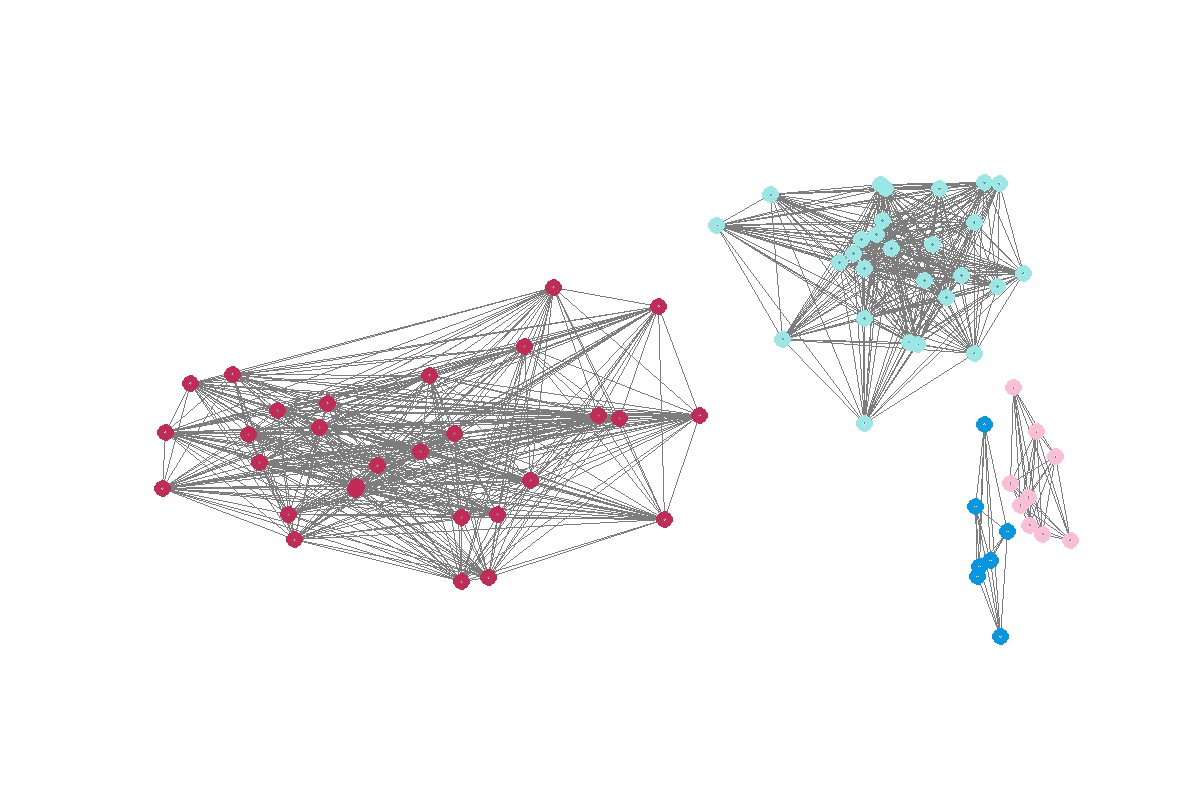}
    }
    \caption{(a): filtered proposals in one image;
             (b): normalized cut proposals in one image;
             (c): filtered nodes in one graph;
             (d): normalized cut nodes in one graph}
    \label{graph_cut}
\end{figure*}

\subsection{Identical Normalization}
The feature of the identical shape of input and output make RepGN easily to be integrated into an existing object detection model. Furthermore, for the reusing of pretrained weights and simplicity of training, we add an identical connection and identical normalization operation to RepGN as Eq.4 depicted:
\begin{equation}
V' = \frac{(\lambda * RepGN(V) + V') - E[V]}{Var[V]+\epsilon}
\end{equation}
where $\lambda$ is a coefficient to control the influence of relation and contextual information and $\epsilon$ exits for numerical stability. Neglecting the $\epsilon$, the refined, local and global contextual-awared representation of proposal $V'$, has same mean and variants with original representation without relational information. This norm operation can weak distrubance to the original detection model caused by RepGN.

\section{Experiments}
\label{sec:exp}

\subsection{Datasets}

 We verify the validity of the proposed method on COCO detection datasets \cite{coco_dataset} with 80 categories, including 118k images for training and 5k for validation. 
Default detection evaluation metrics for COCO are used here, mainly for AP at IoU=.50:.95.\par
To present the effect of graph neural network general on object detection task, we take Faster RCNN \cite{faster-rcnn} and Cascade RCNN \cite{cascade-rcnn} as baseline.
ResNet-50 \cite{resnet} equipped with FPN \cite{fpn} layers is selected as the default backbone for most of experiments, and others are based on ResNet-101 to show the generality.
We implemented the baselines mentioned above and get comparable performance reported in their original papers. Due to the shape input and output of our RepGN module is identical, it's convinent to insert RepGN, after regions proposal network and into the bounding box regressioner and classifer($bbox\_head$).\par

On the training stage, we select SGD as optimizer with weight\_decay = 0.0001, adjusting learning rate dynamically at epoch 8 and 10. 
All models are trained for 12 epochs on NVIDIA Tesla V100 GPUs.
Especially, for different number of GPUs, learning rate need to be set proportional to the number of GPUs, for example, $lr=0.01$ or $lr=0.02$ for 4/8 GPUs respectively.

\subsection{Experiments Setup}

In the Faster-RCNN and other baseline frameworks, RPN generates a $256*7*7$ feature map for each proposal. Next, the $bbox\_head$ converts the feature map to $1024$-dim embedding vector. Two fully-connected layers generate the final logits for classification and regression. Additional, we treat the $(x_1, y_1, x_2, y_2, \frac{x_1+x_2}{2}, \frac{y_1+y_2}{2}, \frac{|x_1-x_2|}{|y_1-y_2|})$ where (x1, y1, x2, y2) are the top-left and bottom-right normalized coordinates of the proposal, as a spatial representation explicitly.\par

In this section, we adopt gradual experiments to investigate whether the information from neighbor proposals and global context, including sparial and sementic representation can boost the classification and regression performance or not.

Firstly, we select a conservative method: keeping the baseline framework and adding a graph convolution network sequence paralleled with shared fully connected layers in $bbox\_head$, aiming to get clear the impact of spatial relationship in object detection.
Spatial information of proposals consists of a concatenation of location and size of each one, which is also regarded as the input vector of vertex. We employ two layers multi-head graph attention networks \cite{petar2018gat} with 8 heads and the output dimension is 1024 aligned with baseline. \par

After that, we attempt to fuse the feature of every proposal according to their relationship information, including spatial and semantic. And RepGN module can be inserted into the pipeline detection after the RPN or into the $bbox\_head$. To evaluate these two methods seperately, we note the first one RepGN(a) and second one RepGN(b). Fruthermore, we conduct experiments on the RepGN module without GCPool branch to investigate the influence of global contextual information on final performance.

\subsection{Experiment Result}

\begin{table}
    \centering
    \begin{tabular}{p{3.1cm}|p{2.4cm}|p{2.0cm}} 
        \hline
        \hline
        Base-Model & Method & $AP_{0.50:0.95}$   \\ 
        \hline
        \multirow{2}{*}{Faster-RCNN 50} & None & $0.364$  \\
        & RepGN   & $\mathbf{0.366}$ \\
        \hline
        \multirow{2}{*}{Faster-RCNN 101} & None & $0.386$  \\
        & RepGN & $\mathbf{0.387}$  \\
        \hline
    \end{tabular}
    \label{s_accuracy}
    \caption{Comparison between baseline and GAT.}
\end{table}

\begin{table}
    \centering
    \begin{tabular}{p{6.2cm}|p{1.575cm}|p{1.575cm}} 
    \hline
    \hline
    Time & ResNet-50 & ResNet-101  \\
    \hline
    Train Time(shared-fcs) & $0.450$ & $0.487$  \\
    Train Time(with GAT layers) & $0.430$ & $0.501$  \\
    Inference Time(shared-fcs) & $10.4$ & $9.5$ \\
    Inference Time(with GAT layers) & $10.2$ & $9.4$ \\
    \hline
    \hline
    \end{tabular}
    \caption{The time consumption of different backbone on $Faster-RCNN$. For the train time, it is $s/iter$. For the inference time, it is $sample/s$. }
    \label{time}
\end{table}
\begin{table}
    \centering
    
    \begin{tabular}{p{3.4cm}|p{4.8cm}|p{1.4cm}} 
        \hline
        \hline
        Base-Model & Method & $AP_{0.50:0.95}$   \\ 
        \hline
        \multirow{4}{*}{Faster-RCNN 50} 
        & None & $0.364$  \\
        & RepGN(a)   & $0.368$ \\
        & RepGN(a)+GCPOOl  & $0.374$ \\
        & RepGN(a)+RepGN(b)+GCPool  & $\mathbf{0.375}$ \\
        \hline
        \multirow{4}{*}{Faster-RCNN 101}
        & None & $0.386$  \\
        & RepGN(a)   & $0.389$ \\
        & RepGN(a)+GCPOOl  & $0.394$ \\
        & RepGN(a)+RepGN(b)+GCPool  & $\mathbf{0.394}$ \\
        \hline
        \multirow{4}{*}{Cascade-RCNN 50} 
        & None & $0.402$  \\
        & RepGN(a)   & $0.402$ \\
        & RepGN(a)+GCPOOl  & $0.407$ \\
        & RepGN(a)+RepGN(b)+GCPool  & $\mathbf{0.407}$ \\
        \hline
        \multirow{4}{*}{Cascade-RCNN 101} 
        & None & $0.420$  \\
        & RepGN(a)   & $0.418$ \\
        & RepGN(a)+GCPOOl  & $0.420$ \\
        & RepGN(a)+RepGN(b)+GCPool  & $\mathbf{0.422}$ \\
        \hline
    \end{tabular}
    \caption{Comparison between baseline and our RepGN module}
    \label{tab_sb}
\end{table}

Table 1 compares the performance of GAT paralleled $bbox\_head$ with original $bbox\_head$, proves the feasibliliy of graph network based methods to enrich the representation of proposals.\par 
Shown in Table \ref{tab_sb}, our RepGN module can  improve the mAP accuracy campared with original pipeline of detection or GAT intergrated method. In detailed, GCPool provide the most remarkable improvement, these results prove the effectiveness of the local and global context information extraction on object detection task. \par
Furthermore, Table \ref{time} indicates that the RepGN can make an significant improvement in AP performance without significant drop in time consumption. The low computation cost makes our RepGN module is compratiable to other relation and context modeling methods in object detection field.

\section{Conclusion}
In this work, we devise our relational proposal graph network(RepGN) architecture to enrich the feature and embedding of or proposals for object detection task. By organizing the proposals to a graph in the spatial domain, we can propagate the information through the graph proposal by using a graph network. In this manner, we can get a better representation of proposals by incorporating both semantic and spatial object relationships. In detailed, we propose a graph pooling method to retrieve a high-level description of all proposals for getting global contextual information in an efficiency way. We evaluate our model on COCO dataset and achieve the stable improvement compared with baseline. 
\label{sec:con}

\bibliographystyle{unsrt}
\bibliography{main}
\end{document}